\documentclass{article}

\usepackage{microtype}
\usepackage{graphicx}
\usepackage{subfigure}
\usepackage{booktabs} % for professional tables

\usepackage[bookmarks=false]{hyperref}

\usepackage[accepted]{icml2020}

\icmltitlerunning{Reinforcement Communication Learning in Different Social Network Structures}

\begin{document}

\twocolumn[
\icmltitle{Reinforcement Communication Learning in Different Social Network Structures}



\icmlsetsymbol{equal}{*}

\begin{icmlauthorlist}
\icmlauthor{Marina Dubova}{iu}
\icmlauthor{Arseny Moskvichev}{uci}
\icmlauthor{Robert L. Goldstone}{iu,iu2}
%\icmlauthor{Aeiau Zzzz}{equal,to}
%\icmlauthor{Bauiu C.~Yyyy}{equal,to,goo}
%\icmlauthor{Cieua Vvvvv}{goo}
%\icmlauthor{Iaesut Saoeu}{ed}
%\icmlauthor{Fiuea Rrrr}{to}
%\icmlauthor{Tateu H.~Yasehe}{ed,to,goo}
%\icmlauthor{Aaoeu Iasoh}{goo}
%\icmlauthor{Buiui Eueu}{ed}
%\icmlauthor{Aeuia Zzzz}{ed}
%\icmlauthor{Bieea C.~Yyyy}{to,goo}
%\icmlauthor{Teoau Xxxx}{ed}
%\icmlauthor{Eee Pppp}{ed}
\end{icmlauthorlist}

\icmlaffiliation{iu}{Cognitive Science Program, Indiana University, Bloomington, IN, USA}
\icmlaffiliation{iu2}{Department of Psychological and Brain Sciences, Indiana University, Bloomington, IN, USA}
\icmlaffiliation{uci}{Department of Cognitive Sciences, University of California, Irvine, CA, USA}
%\icmlaffiliation{to}{Department of Computation, University of Torontoland, Torontoland, Canada}
%\icmlaffiliation{goo}{Googol ShallowMind, New London, Michigan, USA}
%\icmlaffiliation{ed}{School of Computation, University of Edenborrow, Edenborrow, United Kingdom}

\icmlcorrespondingauthor{Marina Dubova}{mdubova@iu.edu}
%\icmlcorrespondingauthor{Eee Pppp}{ep@eden.co.uk}

\icmlkeywords{Multi-Agent Reinforcement Learning, Decentralized Optimization, Language Evolution}

\vskip 0.3in
]



\printAffiliationsAndNotice{}  % leave blank if no need to mention equal contribution

\begin{abstract}
Social network structure is one of the key determinants of human language evolution. Previous work has shown that the network of social interactions shapes decentralized learning in human groups, leading to the emergence of different kinds of communicative conventions. We examined the effects of social network organization on the properties of communication systems emerging in decentralized, multi-agent reinforcement learning communities. We found that the global connectivity of a social network drives the convergence of populations on shared and symmetric communication systems, preventing the agents from forming many local ``dialects". Moreover, the agent's degree is inversely related to the consistency of its use of communicative conventions. These results show the importance of the basic properties of social network structure on reinforcement communication learning and suggest a new interpretation of findings on human convergence on word conventions.% among interacting humans.

%We found that the proportion of global connections in a social network determines the convergence of populations on shared and symmetric communication systems

\end{abstract}

\section{Introduction}
Human languages evolve as complex adaptive systems, driven by micro-level processes and constraints (such as individual learning mechanisms and perceptual biases), macro-level factors (such as a topology of social interactions), and the history of their development \cite{steels2000language, christiansen2008language, five2009language}. %Multi-Agent Reinforcement Learning (MARL) setting is a promising benchmark for simulating these processes as it maintains the flexibility of individual learners, while allowing to scale the simulation to relatively big populations. Human language evolution, convention formation, and decentralized group learning studies, in turn, can suggest the solutions to the optimization problems that arise in the MARLC.

Linguistic communication depends on the shared knowledge of word-to-meaning mapping conventions \citep{lewis1975languages}, upon which the population converges through the local interactions between agents, often with no central controller available \cite{baronchelli2018emergence}. Empirical studies of human learning demonstrate that groups quickly converge on new communicative conventions in ``decentralized" settings %, even when prevented from using natural languages 
\cite{garrod1994conversation, selten2007emergence}.

Multi-agent reinforcement learning to communicate (MARLC), however, faces instability challenges if no central optimization is introduced \cite{bernstein2002complexity, laurent2011world, matignon2012independent}. This influences the ability of groups consisting of reinforcement learners to converge on efficient and stable communication systems which are shared by all their members. Therefore, different methods of centralized control and optimization have been proposed to stabilize MARLC \cite{sukhbaatar2016learning, Foerster_Assael_de_Freitas_Whiteson_2016, lowe2017multi, Pesce_Montana_2020, Foerster_Chen_Al-Shedivat_Whiteson_Abbeel_Mordatch_2018}. Central optimization makes the simulations more brittle and less flexibly adaptive, and, potentially, less promising in developing communication systems as freely expressive and well-optimized for their users \cite{gibson2019efficiency} as natural languages. 

We argue that empirical evidence on individual- and population-level factors that drive decentralized learning in human groups can guide simulations of language evolution in MARLC settings. In this work, we explore whether the social network organization shapes the properties of communication systems that arise through decentralized MARLC in simplified settings. 

\begin{figure}[b]
\centering
\includegraphics[scale=0.20]{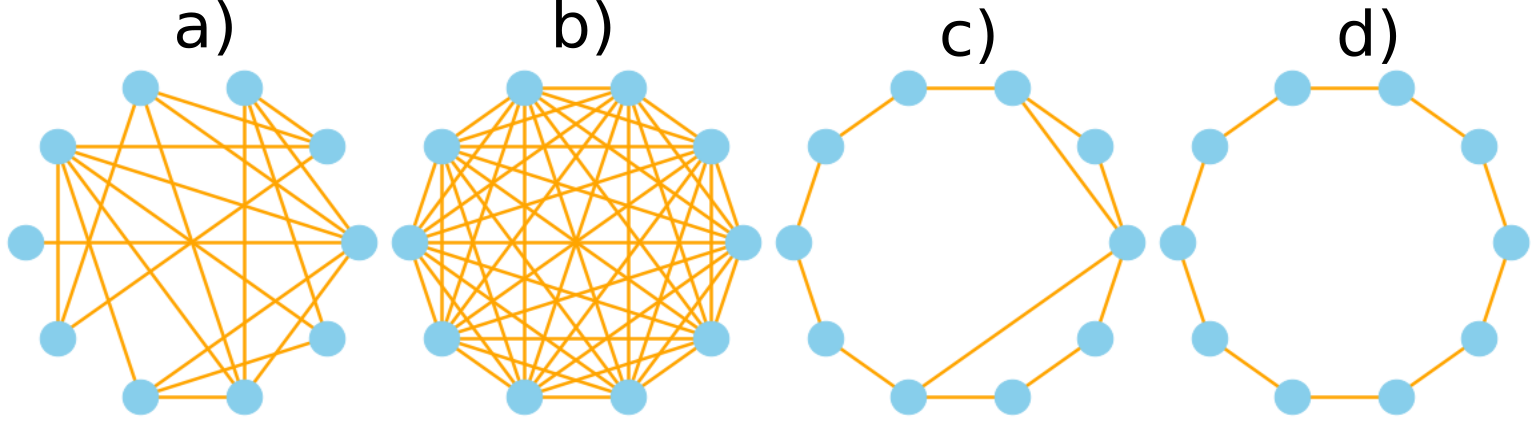}
\caption{Types of social topologies tested in our experiments.\\ a) Random \cite{erdos1959random}, b) Fully-connected (clique), c) Small-world \cite{newman1999renormalization}, d) Ring}
\label{fig:thefigure1}
\end{figure}

\begin{figure*}[h!]
\centering
\includegraphics[scale=0.5]{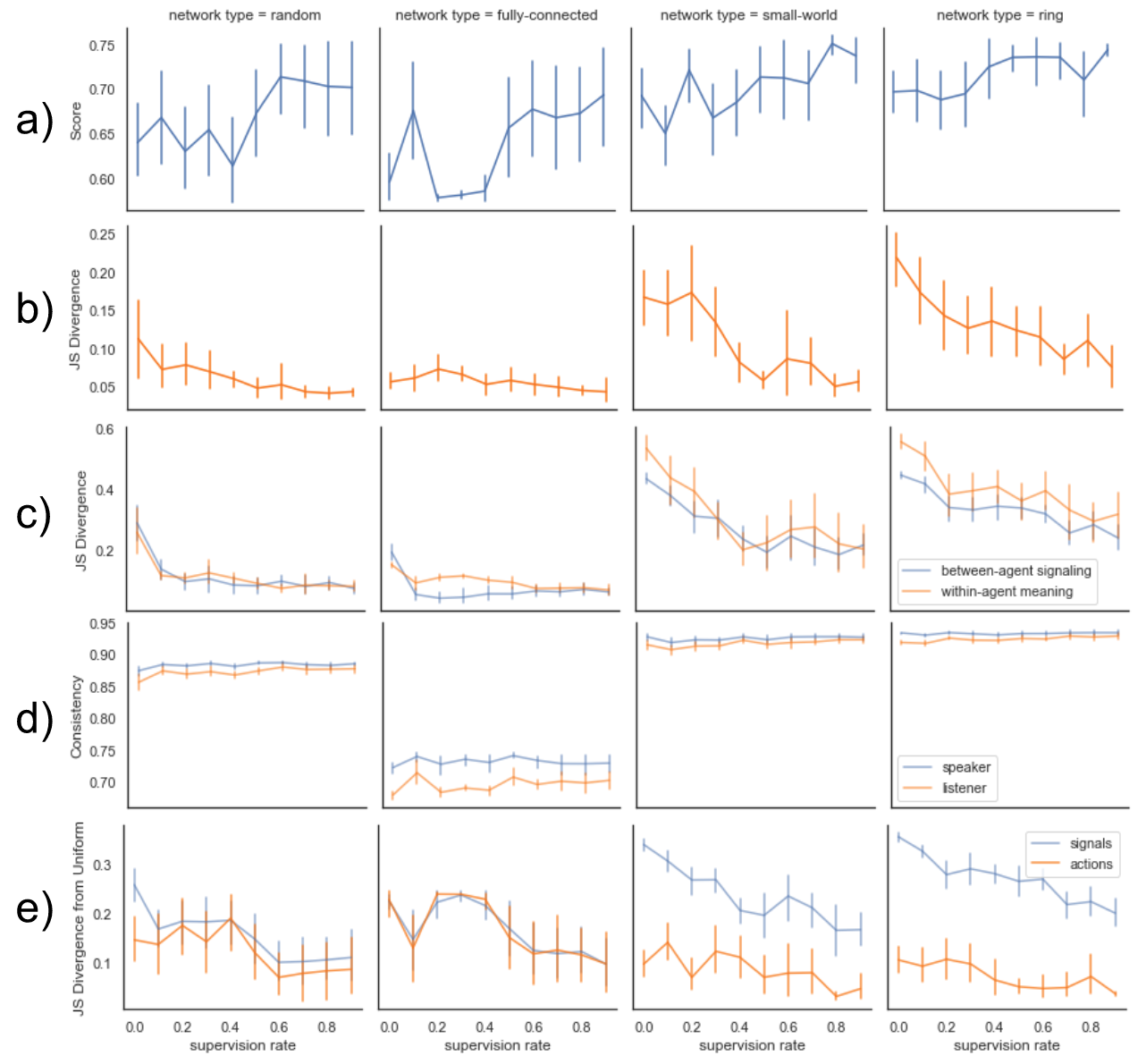}
\caption{The Communication Analysis Metrics with 95\% Confidence Intervals for Experiment 1. a) Average rewards b) Between-agent signal-action mapping divergence c) Signaling divergence (blue) and within-agent signal-action mapping divergence (orange) d) Average speaking (blue) and listening (orange) consistencies e) Average predictability of agents’ signals (blue) and actions (orange).}
\label{fig:thefigure}
\end{figure*}

\subsection{Human Learning in Different Social Network Structures}

Convergence of human groups on word conventions is dramatically affected by the social topology that determines the possible interactions between participants, as demonstrated by the naming game experiment %on large groups of participants
\cite{centola2015spontaneous}. In particular, when arranged in a social network with many local connections (e.g. ``ring'' topology (Fig. 1d)) or a randomly-connected network (Fig. 1a), large groups converge on many local word conventions, reaching no global consensus. However, if each person is equally likely to interact with any other person in the group (``clique'' (Fig. 1b)), global consensus is easily achieved with no centralization. Other studies on decentralized problem solving in human groups demonstrated that social network organization shapes the multi-agent optimization process, with different network types being beneficial for different types of optimization landscapes. High local connectivity supports independent local exploration, whereas high global connectivity helps groups to converge on a shared solution, choosing among the local ones \cite{fang2010balancing, mason2008propagation, mason2012collaborative, lazer2007network, wisdom2011innovation}.

In this study, we looked at how the type of social network organization, its average degree, and local connectivity affect the results of communication learning in groups of deep reinforcement learning agents. %In this work, we looked at whether the social network topology shapes the properties of communication systems that arise through decentralized MARLC in simplified settings. In particular, we tested effects of the type of social topology, average degree of the agents, and proportion of global connections in a social network. 

\section{Method}

\subsection{Coordination Game}

Every game episode involves two agents, randomly assigned to speaker and listener roles. The speaker produces a message, which is transmitted to the listener. Then, both agents independently choose an action\footnote{Action and signaling sizes were set to 4 in all the simulations.}. If the actions match, the agents receive a reward. We add an additional penalty for overusing any specific action to avoid degenerate solutions that ignore the communication channel. %altogether. % and an action, while the listener receives the message and outputs an action
This setting encapsulates the most basic form of a coordination game that benefits from the formation of communicative conventions. Please, see supplementary materials for more detail.

\subsection{Types of Social Network Organization}
Social network determines how the agents are sampled to play games with one another. For each game round, one agent is selected randomly, and then its partner is selected from its neighbors. Thus, an agent can only play a game with one of its immediate neighbors in the social network. In all the simulations, the social network size was set to 10. The networks were undirected, and the self-connections were not allowed. We tested 4 types of social networks in our experiments (Fig. 1): \\
\textbf{1. Random (ER).} Random graph is generated by connecting the nodes with equal probability $p$ \cite{erdos1959random}. \\
\textbf{2. Fully-connected (clique).} In clique, all the nodes are connected to each other. \\
\textbf{3. Ring.} In the ring network, all the nodes have exactly two neighbors, and connections form a single continuous path. %through all the nodes. 
\\
\textbf{4. Small-world.} The small-world network is generated by adding new, ``global", connections to the ring network with constant probability $p$ \citep{newman1999renormalization}.

\subsection{Agents}

We used simple feed-forward neural networks to represent the agents. The networks were trained using a vanilla deep Q-learning algorithm \cite{mnih2015human} with an added proportion of bottom-up driven ``supervising" feedback (see supplementary materials). We tried to avoid any centralized optimization, popular in MARLC settings, to study the ability of different networks to self-organize conventions.

\subsection{Metrics}

We used a number of information-theoretic metrics, developed in \citet{dubovamoskvichev} and \citet{lowe2019pitfalls} to comprehensively evaluate the communication protocols: speaker \& listener consistency, between- and within-agent signal-action mapping divergence, signaling divergence, and behavioral predictability (see supp. materials).

\section{Results}

\subsection{Experiment 1: types of social network organization}
We simulated multi-agent learning in 4 types of social networks: \textbf{ring} (avg degree=2, var=0), \textbf{random} (avg degree=2, var=1.6, p(connection)=0.2), \textbf{small-world} (avg degree=2.2, p(add global connection)=0.2), and \textbf{clique} (avg degree=9, var=0). We also varied the supervision rate (from 0.0 to 0.9 with a step of 0.1). MARLC in each combination of these two conditions was simulated 10 times to get statistical estimates of the communication metrics. In all the experiments, each simulation consisted of 120000 game rounds.

The average speaker and listener consistencies were highest in the ring and small-world networks, and the lowest in the clique structure. Consistency scores did not vary with the supervision rate (Fig. 2d). These patterns indicated the potential dependency of the consistency scores on a single factor: agent's degree (the clique had the highest possible average degree). This hypothesis is tested in Experiment 2.
%Average degree was the lowest in ring, small-world, and random networks, and the latter had the highest variance in degree distribution

The agents in random and fully-connected social networks developed almost perfectly symmetric and homogeneous communication systems according to all three communication asymmetry metrics (Fig. 2b and 2c). Small-world and ring social networks lead the populations to develop asymmetrical and local communication patterns. Moreover, supervised feedback helped the agents in all topology conditions to develop more shared and symmetric communication systems. We hypothesized that the effect of network type on asymmetry scores is driven by local connectivity of the populations: fully-connected and random networks do not form local communities, whereas the ring-shaped and small-world networks in our simulations mainly consisted of local connections. We tested this hypothesis in Experiment 3.

Lastly, agents in fully-connected and random networks produced less diversified actions, but their action and signaling distributions were much more coordinated than in the ring-shaped and small-world network structures (see Fig. \ref{fig:thefigure}). It suggests that agents in the last two conditions overfitted to the ``action diversity" part of the reward function, while failing to coordinate using the communication channel. 

\subsection{Experiment 2: average degree}
To test the effect of network's average degree on speaker and listener consistencies, we focused on the random (ER) network, which produces desired variation in the average degrees of the nodes, while keeping most of the other network properties constant. We varied the probability of connecting nodes in the network from 0.2 to 0.9, and supervision rate from 0 to 0.9 with a step of 0.1. This resulted in 80 conditions total, each of which was simulated 5 times to obtain consistency estimates with confidence intervals.

We found that the average degree negatively affected listener ($p<0.001$) and speaker ($p<0.001$) consistencies (Fig. 3). Consistency scores for the average degrees higher than 5.0 stayed similar to the estimate we obtained for the fully-connected topology in Experiment 1. %Consistency scores stopped decreasing for average degrees higher than 5.0 and stayed similar to the value we obtained for the fully-connected topology in Experiment 1.

\begin{figure}[t]
\centering
\includegraphics[scale=0.27]{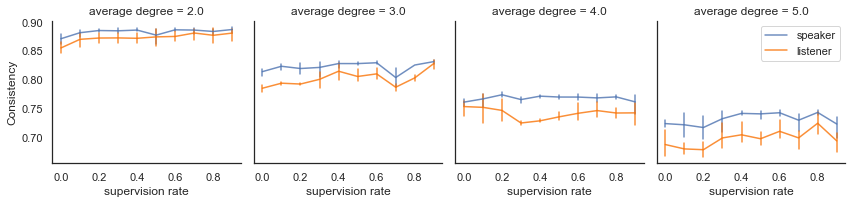}
\caption{Speaker and listener consistency estimates with 95\% Confidence Intervals and the average degree of agents in the random social network (Experiment 2)}
\label{fig:thefigure1}
\end{figure}

\subsection{Experiment 3: global and local connections}
We examined the effect of locally-connected groups in a social network on homogeneity and symmetry of the developed communication systems. For this, we tested MARLC in the small-world network with different probabilities of adding a ``global" connection to the initial ring-shaped structure. We classify a link as ``global" if it connects agents further than one link apart in the ring network. We varied both the probability of global connection and the supervision rate from 0 to 0.9 with a step of 0.1. Each combination of these two factors was simulated 5 times.

Global connection probability was inversely related to the signaling divergence ($p<0.001$), between- ($p<0.001$) and within-agent signal-action mapping divergence ($p<0.001$) of the developed communication systems (Fig. 3). Higher proportion of global connections led the groups to converge on communication systems that are shared by all their agents. Supervised feedback also helped the agents to develop homogeneous and symmetric communication patterns. The cumulative effect of both high supervision rate and probability of global connections lead to communication systems that are shared by all the agents in a group.
\begin{figure}[h]
\centering
\includegraphics[scale=0.32]{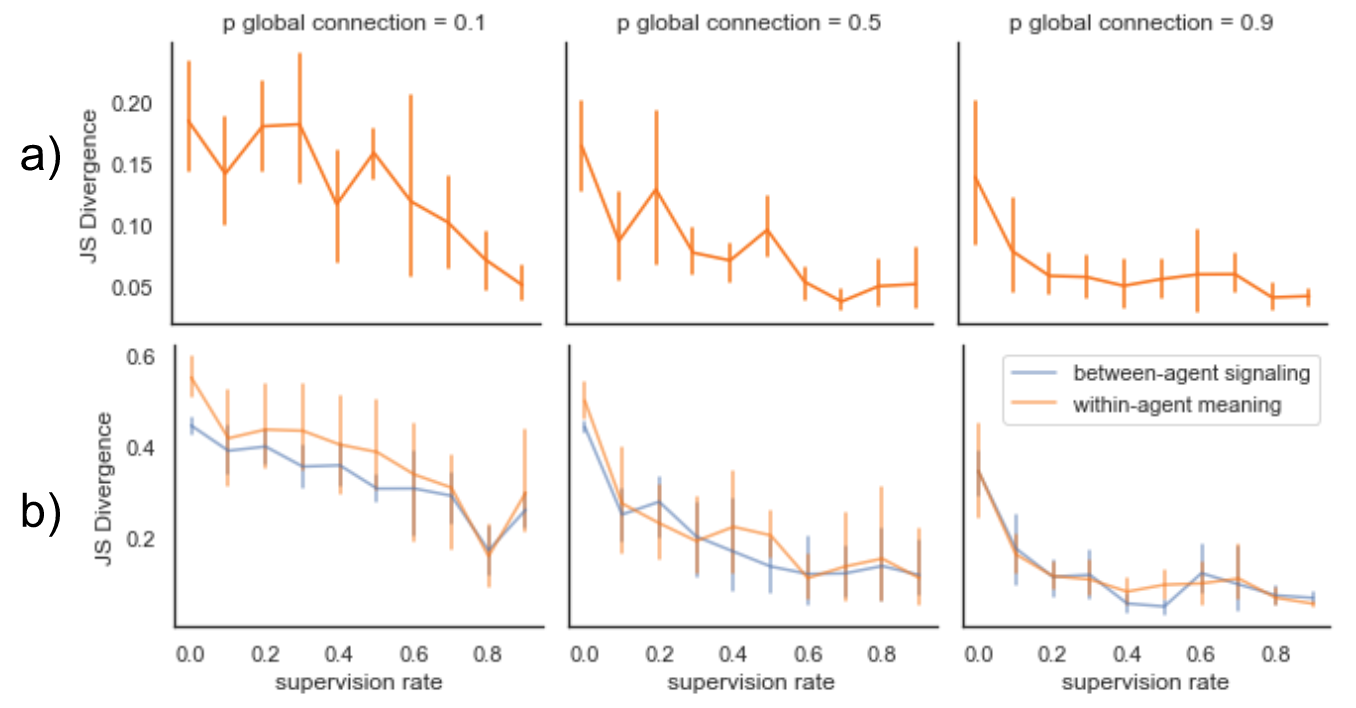}
\caption{The Communication Analysis Metrics with 95\% Confidence Intervals for the Experiment 3. a) Between-agent signal-action mapping divergence b) Signaling divergence (blue), within-agent signal-action mapping divergence (orange) and the probability of global connection in the small-world network}
\label{fig:thefigure2}
\end{figure}

\section{Discussion and Conclusion}

We conducted three experiments to explore and isolate the factors of social network organization that drive the effectiveness, homogeneity, and symmetry of communication systems developed in MARLC settings. Following the suggestions of \citet{lowe2019pitfalls}, we used a set of information-theoretic metrics to evaluate our results. This allowed us to determine which particular properties of communication systems are affected by our interventions.

The results of our first experiment partially replicated the effects of social network type on learning word conventions by human groups \cite{centola2015spontaneous}. This suggests that the simple domain-general reinforcement learning model can capture core regularities found in human convergence on linguistic conventions. Our further experiments provide a more detailed account of particular social network properties that are responsible for these patterns in communication learning. %To our knowledge, it is the first time that the average degree and global connection probability were tested to explain the results of multi-agent (human and machine) learning to coordinate through communication.

The second experiment demonstrated that the average degree of a social network is a central factor affecting how deterministically agents use the communication channel. The more communication partners an agent needs to adapt to, the more variable are its signaling and listening patterns. Key finding of the third experiment is that the proportion of global connections is the primary determining factor for three variables: homogeneity, within- and between-agent symmetry of the developed communication system. In particular, high proportion of local social connections led to the emergence of local communication patterns (``dialects") within the population, whereas adding more global connections forced the agents to find global consensus. Similarly, high local connectivity resulted in asymmetric communicative patterns, where even the same agent in two different roles uses completely different vocabularies. 

At present, social network effects on MARLC have been largely under-investigated. To the best of our knowledge, there is only one work in this direction; L. Graesser and her colleagues (2019) looked at how communication systems developed within communities change if one community is introduced to another, depending on their inter- and intra-connectivity. Unfortunately, the insights from that work are only applicable for the specific scenario of community merging. In our study, we focused on the social network factors that are applicable to almost any MARLC setting.

The results of our analysis corroborate Lowe et al.'s (2019) suggestion that overall scores do not reflect whether the agents use the communication channel to solve the task, and how efficiently they do so (see Fig. 2). For example, the performance scores of the agents learning to coordinate in a ring-shaped social network were the highest among all the network types. However, more detailed analysis revealed that agents in this condition developed many local communication patterns, which varied across agents and their roles. This, again, illustrates the importance of thorough analysis of communication protocols learned in MARLC settings. 

Our approach has a number of limitations that are important to mention. Firstly, while we aimed to test a broad spectrum of social network factors, our experiments are by no means comprehensive. There are other important aspects of social network structure that may play a key role in determining the properties of emerging communication protocols. We believe that studying the effects of variance of network's degree distribution and its modularity on MARLC is especially promising. Secondly, we used a very simplified setting of vanilla Q-learning in an ``amodal" coordination game to minimize the number of assumptions that might make our results less representative to the MARLC problem in general. We suggest testing these results on more advanced reinforcement learning models and realistically perceptually grounded game settings.

\section{Data Availability}
The code and data for this work are available online at the \href{https://github.com/blinodelka/Multiagent-Communication-Learing-in-Networks}{project's github repository}. 

\section{Acknowledgements}
Authors thank Yong-Yeol Ahn, Andrei Amatuni, Thomas Gorman, Jack Avery, Ben Kovitz, all the attendees of the PCL lab meetings, and two anonymous reviewers for their valuable feedback on this project. This research was supported in part by Lilly Endowment, Inc., through its support for the Indiana University Pervasive Technology Institute.
https://kb.iu.edu/d/anwt\#carbonate

\bibliography{bibliography}
\bibliographystyle{icml2020}
%\pagebreak
%\newpage
\appendix

\section{Coordination game environment}

Each game consists of two-step episodes. Every episode of the game involves two agents: a speaker and a listener who need to coordinate their actions. 

On the first step, the speaker produces an action and a message. On the second step, the listener receives the message and outputs an action. If the actions match, agents receive a fixed ``coordination'' reward $R_c$. There are four unique actions and messages that the agents choose from.

In order to avoid trivial solutions when agents converge on a single action and ignore the communication channel, we introduced a penalty for repeating an action too often. Specifically, an additional reward of $min(0, 1/4 - \hat{p}_A)$ is added when an agent chooses action $A$. Here $\hat{p}_A$ denotes an empirical proportion of selecting action $A$ during the last $H$ steps. We used a fixed length history $H=100$. Current $\hat{p}_A$ are added to the agent's input state spaces to aid convergence.

On every episode, each agent can be assigned to play either the speaker or the listener role. The current role type is provided as a binary input.

Overall, the agents receive, as inputs:
\begin{enumerate}
    \item A binary role indicator
    \item Random input (for speakers, in order to allow for non-deterministic policies) or speaker's message (for listeners)  
    \item Proportions of different actions in agent's recent history
\end{enumerate}

The agents have two output layers both stemming from the last hidden layer:

\begin{enumerate}
\item Action output layer (for neurons, and the action is determined as argmax)
\item Message output layer which has the same structure as the action output layer. Message outputs are ignored if the agents plays the listener role
\end{enumerate}

\section{Agents}

Agents are implemented using simple feed-forward neural networks with two hidden layers (hidden sizes of 25 and 15) and ReLU activation function.

We used vanilla deep Q-learning algorithm with additional supervised updates to train the agents. Supervised feedback corresponds to "peeking" the other agent's action and storing it in memory (with the signaling experience from that game) as a "correct" response. In this case, the action-signal mapping is not superimposed, and the "correct" answers in the supervising trials are bottom-up driven and reflect the dynamics of the agents themselves. Supervised feedback was implemented by changing a certain proportion of negative (``miscoordinated") experiences to the ``supervising" ones. We decided to control for supervision in our experiments because partial supervising feedback in different forms is often available in naturalistic language learning situations. By amplifying the reinforcement signal, supervising feedback may help to overcome the ``sparcity of rewards" problem in multi-agent reinforcement learning.

\section{Metrics}

We follow the same set of metrics as used in \cite{dubovamoskvichev}.

%For example, we may need $p_i^s(a, m)$ which denotes a joint distribution over actions and messages for agent $i$, when agent $i$ plays the speaker role. 

%The analysis was conducted for the last 4000 games of each simulation, and the metrics were averaged for all the agents, and then for each simulation. The obtained values were accumulated for each condition for getting statistical estimates and confidence intervals.

\begin{enumerate}
    \item \textbf{Speaking Consistency and Listening Consistency}.
    
    This metric provides a quantitative measure of whether the actions that the agents perform are related to the signals that they send or receive. We follow \citet{lowe2019pitfalls} who proposed to use the normalized mutual information between the distributions over messages and actions induced by an agent. The metric can be formally described as follows:
    
  % This metric has a number of intuitive properties that ease its interpretation. Firstly, it is bound between $0$ and $1$, where $1$ corresponds to a situation when an action is fully predictable from the messages. If the agent chooses an action independently of the messages it receives or sends, the (normalized) mutual information is equal to zero. Secondly, the metric does not impose any constraints on the form through which the messages and actions can be related. 
    \begin{equation}\label{f1}
        C = \sum_{a \in A_l} \sum_{m \in A_c} p_{a, m}(a, m) log \frac{p_{a,m}(a,m)}{p_a(a)p_m(m)} / Z
    \end{equation}
    
    Here, $Z$ is the average entropy of the two marginal distributions: $Z = \frac{(H(p_a) + H(p_m))}{2}$. $A_l$ and $A_c$ denote the set of available actions and the set of available messages respectively.
    
   This metric is computed twice for every agent, conditioned on the role played by the agent (speaker or listener). We average these metrics across agents in every simulation to obtain the speaking and listening consistency metrics that we report.
    
    \item \textbf{Communication Asymmetry metrics}:\\
    \textbf{a.} \textbf{Between-agent signal-action mapping divergence}. 
    While the previous metric aimed to measure whether an agent is using communication channel (i.e. whether its actions correspond to the signals in any way), the second metric aims to measure whether the communication patterns differ between agents. 
    
    We compute the average Jensen-Shannon pairwise divergence between distributions of agents' actions following a specific signal (averaged over all signals and pairs of agents).
    
    For a pair of agents, the metric is defined as follows
    \begin{equation}\label{metric2}
         \sum_{m \in A_c} JSD(p_{a_1|m}, p_{a_2|m}) / |A_c|
    \end{equation}
    
    Here, $p_{(a_1|m)}$ are action distributions of agent 1 conditional on the message (received or sent) being equal to $m$. %We average this metric separately across all pairs of speakers and across all pairs of listeners to obtain the speaker and listener between-agent signal-action mapping divergences.
    
    \textbf{b.} \textbf{Within-agent signal-action mapping divergence}. 
    
    %While the previous metric aimed to capture the differences in languages that different agents use. 
    This metric aims to capture internal inconsistencies in agent's behaviors when it switches between roles: whether the agent's behaviors differ depending on whether it receive or send a specific message.
    
    For that, we use the average Jensen-Shannon divergence between the distributions over agent's actions (conditioned on receiving or sending a specific message) when the agent plays the speaker and the listener role. Formally, we use the same definition as in Equation \ref{metric2}, but now $p_{a_1 | m}$ and $p_{a_2 | m}$  correspond to the same agent's distributions when this agent plays different roles (as opposed to distributions of a pair of different agents playing the same role). We average the metric scores across all agents to obtain the final measure that we report.
    
    \textbf{c.} \textbf{Signaling divergence}. 
    
    This metric aims to measure the difference in individual agents' messaging preferences.
    We define \emph{signaling divergence} as an average pairwise Jensen-Shannon divergence of marginal signaling distributions of different agents.
    
    \item \textbf{Behavioral Predictability}. 
    This last metric is created to assess the general diversity in agent's actions. When the agents' actions are less diverse (and hence, more predictable), it is easier to achieve successful multi-agent coordination without using the communication channel. To look at whether the diversity of actions corresponds to the diversity of signals, we also compute the predictability of agents' signaling patterns. We define \emph{behavioural action/message predictability} as Jensen-Shannon divergence between marginal distributions of agent's actions/messages and the uniform distribution.
    
\end{enumerate}

Many of these metrics require knowing probability distributions. We estimate all such distributions empirically.

As a short summary, if we see the learned language as a simple probabilistic dictionary that maps messages to actions, the metrics can be summarized as follows (note that every agent defines two such dictionaries: one for the speaker and one for the listener role):

\begin{enumerate}
    \item \textbf{Speaking Consistency and Listening Consistency}.
    Are the dictionaries reliable? I.e. if we look up a specific message, do we consistently get the same action, or is there a lot of randomness?
     \item \textbf{Communication Asymmetry}:\\
    \textbf{a.} \textbf{Between-agent signal-action mapping divergence}.
    Are the dictionaries similar for different agents?
    
    \textbf{b.} \textbf{Within-agent signal-action mapping divergence}.
    How different are the ``speaker'' and ``listener'' dictionaries that each agent defines?
    
    \textbf{c.} \textbf{Talking divergence}.
    Do different agents show different patterns in their dictionary lookups?
    \item \textbf{Behavioral Predictability}.
    How uniformly do agents look up different words in the dictionary (speaking predictability)? How uniform are the results of their queries (behavioral predictability)?
\end{enumerate}

\section{Statistical analysis}

All hypotheses were tested using a linear regression model with robust covariance estimation, controlling for supervision rate. Excluding the supervision rate did not qualitatively change the results, however.

\end{document}